\documentclass[runningheads]{llncs}

\usepackage{eccv}

\usepackage{eccvabbrv}

\usepackage{graphicx}
\usepackage{booktabs}

\usepackage[accsupp]{axessibility}  

\usepackage{hyperref}

\usepackage{orcidlink}

\usepackage[font=footnotesize, skip=1pt]{caption}
\begin{document}

\title{Bring Music The Horizon: Music-Driven 360$^\circ$ Video Generation}


\author{Kai Hsu, Tsai\inst{1}\orcidlink{0009-0001-7505-6602} \and
Yong Wei, Fu\inst{1}\orcidlink{0009-0000-7709-4437} \and
Hung I, Yang\inst{1}\orcidlink{0009-0000-4554-718X} \and
Yu-Chih Chen\inst{1}\orcidlink{0000-0003-2710-083X}}

\authorrunning{Tsai et al.}

\institute{Department of Computer Science, National Yang Ming Chiao Tung University, Hsinchu 300093, Taiwan
\newline
\email{ \{chris.cs12, willyfu0905.cs12, holdtensec.cs12, berriechen\}@nycu.edu.tw}}

\maketitle
\vspace{-5mm}
\begin{abstract}
Music visualization offers a powerful way to enhance listeners' understanding and experience of music by translating auditory signals into visual forms. However, most existing approaches either rely heavily on lyrics or generate flat, non-immersive videos similar to conventional music videos, which limits their ability to convey the emotional dynamics of music and provide an immersive listening experience.

We propose \textit{Bring Music The Horizon}, an emotion-aware pipeline for music-driven 360$^\circ$ video generation. Given an input song, our work first estimates its emotional trajectory by predicting valence-arousal values~\cite{russell1980circumplex} at the level of every four bars. These values are then converted into emotion-aware visual guidance using EmotiCrafter~\cite{he2025emoticrafter}, and these guidance vectors can be manipulated by the SEGA framework~\cite{brack2023sega}, which provides fine-grained semantic control for keyframe generation. Finally, image-to-video models are applied to the generated keyframes to synthesize temporally continuous 360$^\circ$ videos for immersive music visualization.

Our pipeline generates 360$^\circ$ music visualization videos that reflect the emotional progression and temporal structure of the input song. We demonstrate its capability using songs from different genres and provide qualitative comparisons with \textit{From-Sound-To-Sight}~\cite{vitasovic2025fromsoundtosight}, a representative audio-to-visual generation baseline, on our project page at \url{https://etoile-et-toi-mp3.github.io/BMTH_Project_Page/}.
\vspace{-3mm}
\end{abstract}
\vspace{-5mm}
\section{Introduction}
\label{sec:intro}
\vspace{-2mm}
\subsection{Motivation}
\vspace{-2mm}
Music visualization aims to enhance how listeners perceive and experience music by translating auditory signals into visual content. However, existing approaches are often limited to fixed-view or flat videos, resembling conventional music videos rather than immersive visual experiences. Many methods also rely heavily on lyrics to infer visual semantics, which limits their applicability to instrumental music and may overlook the continuous emotional dynamics embedded in the audio itself.

To address these limitations, we propose \textit{Bring Music The Horizon}, a music-driven pipeline for generating immersive 360$^\circ$ videos. Our method focuses on modeling the temporal evolution of musical emotion and synthesizing 360$^\circ$ visual scenes that align with both the affective and structural progression of the song.

\subsection{Related Work}
\vspace{-1mm}
Recent music visualization methods combine music information retrieval (MIR) and generative models to synthesize visual content from audio. However, many approaches rely on lyrics or other surface-level semantic cues, which may weaken the connection between the generated visuals and the intrinsic musical properties such as rhythm, intensity, timbre, and emotional progression. Moreover, most prior work targets narrative-style flat music videos rather than immersive 360$^\circ$ visualization.

Jeong et al.~\cite{jeong2023tpos} proposed an audio-reactive video generation model that maps audio signals to semantic and temporal visual dynamics, such as generating rain intensity according to audio magnitude. Our work shares the broader goal of audio-reactive visual generation, but focuses specifically on music-driven immersive visualization, where both emotional trajectories and musical structure are used to guide 360$^\circ$ video generation.
\vspace{-1mm}
\subsection{Our Pipeline and Limitations}
\vspace{-1mm}
Our pipeline consists of three main modules. The \emph{(i) MIR module} extracts musical features such as tempo, downbeat timestamps, and valence-arousal (V-A) values~\cite{russell1980circumplex} for each four-bar segment. The \emph{(ii) Emotion-guided 360$^\circ$ Keyframe Generation module} incorporates the estimated emotional features into the visual generation process to produce affect-aligned 360$^\circ$ keyframes. The \emph{(iii) 360$^\circ$ Video Generation module} then applies image-to-video (I2V) models to generate dynamic clips and transition clips, which are concatenated into the final immersive music visualization video.

While the proposed pipeline can generate emotion-aware and temporally structured 360$^\circ$ music visualization videos, several limitations remain. Similar V-A values may produce repetitive keyframes, and visible seams can appear due to boundary inconsistencies in generated 360$^\circ$ videos. The current output resolution is still limited, and the forward-moving visual effect cannot yet be fully controlled, which may affect motion comfort. In addition, overly long prompts may degrade 360$^\circ$ generation and cause the model to produce flat rather than 360$^\circ$, immersive visuals.
\vspace{-3mm}
\section{Method}
\label{sec:method}

\begin{figure}[tb]
  \centering
  \includegraphics[width=1\linewidth]{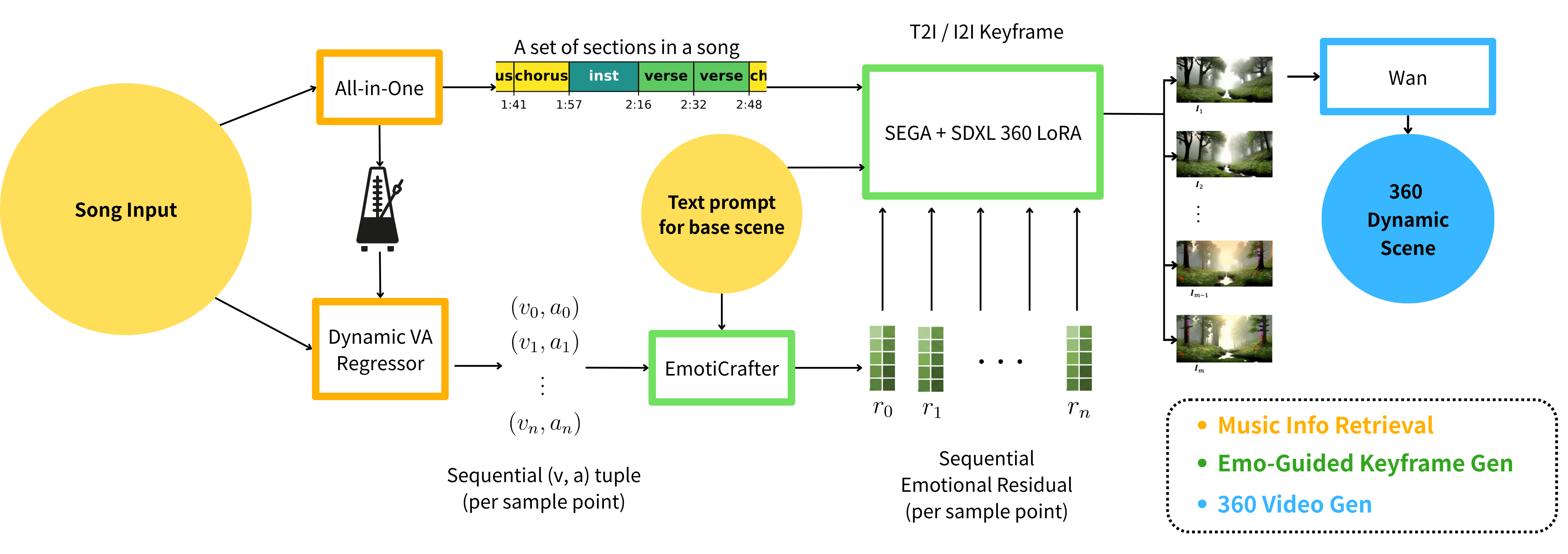}
  \caption{Overview of the proposed music-driven 360$^\circ$ visualization pipeline. Given an input song and a user-specified base prompt, the MIR module extracts musical structure and downbeat-level timing cues, while the Dynamic V-A Regressor estimates a sequence of valence-arousal conditions. These emotion conditions are converted into emotional residuals by EmotiCrafter~\cite{he2025emoticrafter} and used with SEGA~\cite{brack2023sega} and SDXL~\cite{podell2023sdxl} 360$^\circ$ LoRA~\cite{artificialguybr2024redmond360} to generate emotion-aware panoramic keyframes. The keyframes are then animated by Wan-based image-to-video generation to produce the final 360$^\circ$ dynamic visualization.}
  \label{fig:pipeline}
\end{figure}
\vspace{-1mm}
Given an input song $S$ and a user-specified base prompt $p_{\mathrm{base}}$, our pipeline generates an emotionally and temporally aligned 360$^\circ$ music visualization. As shown in Fig.~\ref{fig:pipeline}, the song is first analyzed to extract its bar-level temporal structure and a sequence of valence-arousal (V-A) values that describe the emotional trajectory of the music. These musical cues are then used to adapt the base prompt into emotion-aware panoramic keyframes, allowing the visual content to preserve the user-provided concept while evolving with the song's emotional progression. Finally, the generated keyframes are animated and connected in musical order to produce a complete 360$^\circ$ video whose scene dynamics and transitions follow the structure of the input song.

Formally, the overall process is defined as
\vspace{-1mm}
\begin{equation}
(S, p_{\mathrm{base}}) \rightarrow (B, U, E) \rightarrow K \rightarrow V_{\mathrm{360}}
\end{equation}
\vspace{-1mm}
where $B=\{b_i\}_{i=1}^{N}$ denotes the extracted downbeat timestamps, $U=\{u_t\}_{t=1}^{T}$ denotes the resulting four-bar music units, and $E=\{e_t\}_{t=1}^{T}$ denotes the corresponding emotional trajectory. Each emotion condition is represented as $e_t=(v_t,a_t)$ in the V-A space. The keyframe generation stage produces a sequence of panoramic keyframes $K=\{k_t\}_{t=1}^{T}$, and the video generation stage synthesizes the final 360$^\circ$ music visualization $V_{\mathrm{360}}$.
\vspace{-4mm}
\subsection{Music Information Retrieval (MIR)}
\vspace{-2mm}
The MIR module analyzes the input song $S$ and extracts its temporal structure and emotional trajectory:
\begin{equation}
f_{\mathrm{MIR}}(S) = (B, U, E).
\end{equation}
We first use All-In-One~\cite{kim2023allinone} to estimate downbeat timestamps $B$ and functional segment information, which provide the temporal structure of the song. Based on the extracted downbeats, we group the song into four-bar units $U$. For each unit $u_t$, our proposed Dynamic Valence-Arousal regressor predicts an emotion condition $e_t=(v_t,a_t)$, where $v_t$ and $a_t$ denote the valence and arousal values, respectively. This regressor is trained to estimate segment-level emotional dynamics directly from music audio, rather than relying on lyrics or manually assigned emotion labels. The resulting V-A sequence $E$ is used as the emotional condition for subsequent visual generation, enabling the generated visuals to follow both the musical structure and the emotional progression of the song.
\vspace{-4mm}
\subsection{Emotion-guided 360$^\circ$ Keyframe Generation}
\vspace{-2mm}
The keyframe generation module maps each four-bar music unit to an emotion-aware panoramic keyframe:
\vspace{-1mm}
\begin{equation}
f_{\mathrm{key}}(p_{\mathrm{base}}, e_t) = k_t.
\end{equation}
\vspace{-1mm}
For each unit $u_t$, the base prompt $p_{\mathrm{base}}$ and its corresponding emotion condition $e_t$ are fed into a retrained EmotiCrafter model~\cite{he2025emoticrafter} to obtain an emotional residual $r_t$. This residual describes how the visual semantics of the base prompt should be shifted toward the target emotion. We then use $r_t$ as a SEGA guidance condition $g_t$~\cite{brack2023sega}, enabling fine-grained semantic control during the diffusion process. Finally, SDXL~\cite{podell2023sdxl} with a 360$^\circ$ LoRA adapter~\cite{artificialguybr2024redmond360} is used to generate the panoramic keyframe $k_t$. This design allows each keyframe to maintain the user-specified scene concept while reflecting the emotion predicted from the corresponding music segment. Repeating this process for all $T$ units yields the keyframe sequence $K=\{k_t\}_{t=1}^{T}$.
\vspace{-4mm}
\subsection{360$^\circ$ Video Generation}
\vspace{-2mm}
The video generation module transforms the keyframe sequence $K$ into the final immersive visualization:
\begin{equation}
f_{\mathrm{video}}(K) = V_{\mathrm{360}}.
\end{equation}
Each keyframe $k_t$ anchors one four-bar video unit. The first three bars are synthesized as a dynamic scene $d_t$ using Wan-I2V~\cite{wanteam2025wan}. The final bar is generated as a transition clip $\rho_t$ using Wan-flf2v~\cite{wanteam2025wan}, which connects the current dynamic scene to the next keyframe $k_{t+1}$. By concatenating all dynamic scenes $\{d_t\}_{t=1}^{T}$ and transition clips $\{\rho_t\}_{t=1}^{T}$ in temporal order, the module produces a complete 360$^\circ$ music visualization video $V_{\mathrm{360}}$, in which scene changes are aligned with the temporal structure and emotional flow of the input song.
\vspace{-4mm}
\section{Results and Conclusion}
\label{sec:conclusion}
\vspace{-2mm}
The experimental results show that the proposed pipeline can generate 360$^\circ$ music visualization videos that are directly viewable in VR headsets. The generated scenes follow the temporal structure of the input songs, while the visual atmosphere and scene transitions reflect the estimated emotional trajectory at different musical segments. These results demonstrate the potential of extending conventional music visualization from flat videos to immersive VR experiences.

Overall, this work presents an emotion-aware and structure-aligned pipeline for music-driven 360$^\circ$ video generation. By integrating dynamic music emotion prediction, emotion-guided keyframe generation, and 360$^\circ$ image-to-video synthesis, the proposed method provides a practical framework for immersive music visualization. In addition, retraining EmotiCrafter with a custom dataset helps reduce undesired human-activity artifacts from the original model, resulting in cleaner and more scene-focused visual outputs. We believe this work establishes a promising foundation for future research on cross-modal and immersive content creation.
\vspace{-3mm}
\paragraph{Acknowledgements.}
This work was supported by the MOE Yushan Young Scholar Program under Grant MOE-114-YSFEE-0010-008-P1 and NSTC Taiwan under Grant NSTC 115-2813-C-A49-161-E.

%
%
\bibliographystyle{splncs04unsrt}
\bibliography{main}
\end{document}